\pdfoutput=1

\documentclass[11pt]{article}

\usepackage[final]{coling}

\usepackage{times}
\usepackage{latexsym}
\usepackage{float}
\usepackage{times}
\usepackage{latexsym}
\usepackage{inconsolata}
\usepackage{booktabs}
\usepackage{multirow}
\usepackage{array}
\usepackage{varwidth}
\usepackage{amsmath}
\usepackage[T1]{fontenc}
\newcommand{\yeslogo}{\includegraphics[scale=0.030]{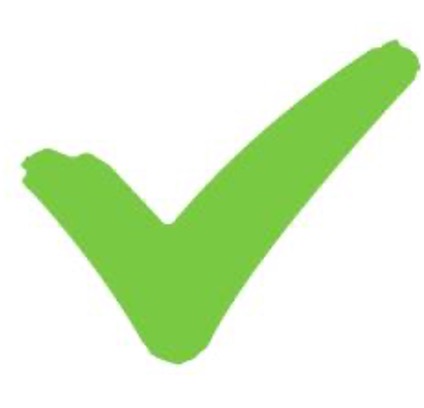}}

\newcommand{\nologo}{\includegraphics[scale=0.030]{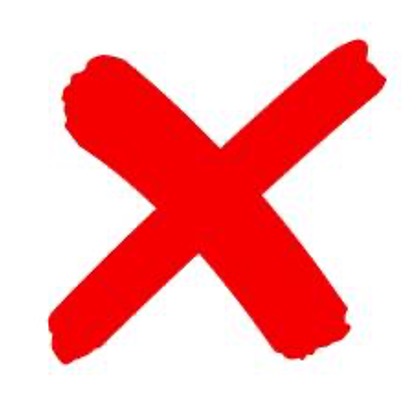}}
\usepackage[utf8]{inputenc}

\usepackage{microtype}

\usepackage{inconsolata}

\usepackage{times}
\usepackage{latexsym}

\usepackage{todonotes}
\usepackage{amsmath}

\usepackage{optidef}
\usepackage{amsfonts}
\usepackage[T1]{fontenc}
\usepackage[per-mode = symbol]{siunitx}
\usepackage[utf8]{inputenc}

\usepackage{microtype}

\usepackage{inconsolata}

\usepackage{booktabs}
\usepackage{multirow}
\usepackage{array}
\usepackage{varwidth}
\newcolumntype{M}{>{\begin{varwidth}{16cm}}l<{\end{varwidth}}} 
\newcolumntype{E}{>{\begin{varwidth}{8cm}}l<{\end{varwidth}}} 
\newcolumntype{N}{>{\begin{varwidth}{5cm}}l<{\end{varwidth}}} 
\newcolumntype{P}[1]{>{\centering\arraybackslash}p{#1}}
\usepackage{soul,color}
\newcommand{\reduce}[1]{#1{\color[HTML]{CB4335}$\dagger$}}

\usepackage{xcolor, soul,todonotes} 
\usepackage{graphicx}
\usepackage{booktabs}
\usepackage{amsmath}
\usepackage{multirow}
\usepackage{array}
\usepackage{varwidth}
\newcolumntype{M}{>{\begin{varwidth}{16cm}}l<{\end{varwidth}}} 
\newcolumntype{E}{>{\begin{varwidth}{8cm}}l<{\end{varwidth}}} 
\newcolumntype{N}{>{\begin{varwidth}{5cm}}l<{\end{varwidth}}} 
\newcolumntype{P}[1]{>{\centering\arraybackslash}p{#1}}
\usepackage{soul,color}
\title{DnA-Eval: Enhancing Large Language Model Evaluation \\through Decomposition and Aggregation}

\newcommand{\salesforce}{$^\ddag$}
\newcommand{\nus}{$^\dagger$}
\newcommand{\astar}{$^\S$}
\newcommand{\ntu}{$^\P$}

\author{Minzhi Li \nus \astar \hspace{1.5em}
        Zhengyuan Liu \astar \hspace{1.5em}
        Shumin Deng \nus \hspace{1.5em}\\
        \bf Shafiq Joty \salesforce \ntu \hspace{1.5em}
        \bf Nancy F. Chen \astar \hspace{1.5em}
        \bf Min-Yen Kan \nus \hspace{1.5em}\\
        \nus National University of Singapore \hspace{0.1em}
        \salesforce Salesforce Research \hspace{0.1em} \ntu Nanyang Technological University\\
        \astar Institute for Infocomm Research (I$^2$R), A*STAR \\
        \texttt{\href{mailto://li.minzhi@u.nus.edu}{li.minzhi@u.nus.edu}} \hspace{1.5em}
        \texttt{\href{mailto://liu\_zhengyuan@i2r.a-star.edu.sg}{liu\_zhengyuan@i2r.a-star.edu.sg}}  \hspace{1.5em}
        \texttt{\href{mailto://shumin@nus.edu.sg}{shumin@nus.edu.sg}}\\
        \texttt{\href{mailto://sjoty@salesforce.com}{sjoty@salesforce.com}}  \hspace{1.5em}
        \texttt{\href{mailto://nfychen@i2r.a-star.edu.sg}{nfychen@i2r.a-star.edu.sg}}  \hspace{1.5em}
        \texttt{\href{mailto://kanmy@comp.nus.edu.sg}{kanmy@comp.nus.edu.sg}} 
}

\begin{document}
\maketitle
\begin{abstract}
The acceleration of Large Language Models (LLMs) research has opened up new possibilities for evaluating generated text. Though LLMs serve as scalable and economical evaluators, how reliable these evaluators is still under-explored. Prior research efforts in the meta-evaluation of LLMs as judges limit the prompting of an LLM to a single use to obtain a final evaluation decision.  
They then compute the agreement between LLMs' outputs and human labels.  
This lacks interpretability in understanding the evaluation capability of LLMs. 
In light of this challenge, we propose \textit{DnA-Eval}, which breaks down the evaluation process into decomposition and aggregation stages based on pedagogical practices.
Our experiments show that it not only provides a more interpretable window for how well LLMs evaluate, but also leads to improvements up to 39.6\% for different LLMs on a variety of meta-evaluation benchmarks.
\end{abstract}

\section{Introduction}

The advancement in Large Language Model (LLM) research has made remarkable progress as LLMs nowadays are able to effectively handle a diverse range of tasks with impressive performance \citep{bang2023multitask}. The capability of LLMs as a general purpose Natural Language Processing (NLP) task solver \citep{qin2023chatgpt} has opened up opportunities in evaluating open-ended text generation tasks \citep{zeng2023evaluating}. On the other hand, the traditional use of human subjects for text evaluation is costly, lacks scalability and reproducibility \citep{karpinska2021perils}. Given LLMs' general capability in NLP tasks and limitations of human evaluation, using LLM-as-a-judge has emerged as an alternative addressing all three issues (cost, scalability and consistency).
\begin{figure}[!ht]
    \centering \includegraphics[width=\linewidth]{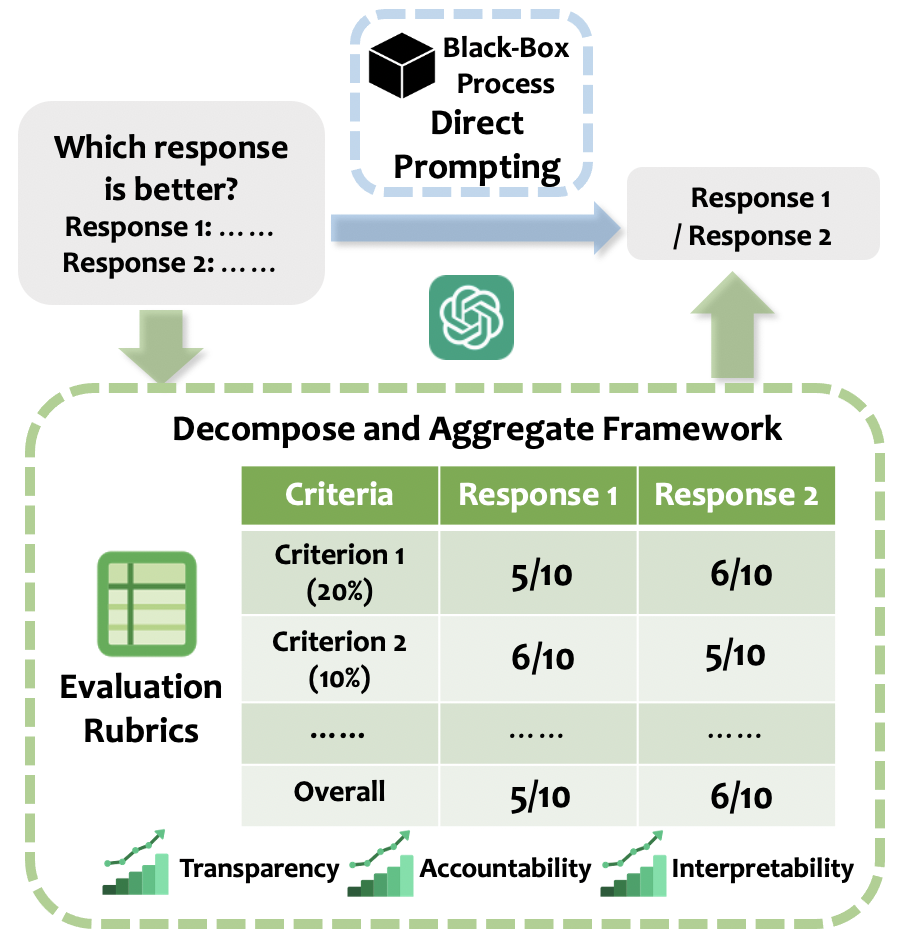}
    \caption{
    \small Different from most previous work which asks LLMs directly for its preference over two responses, our proposed \textit{DnA-Eval} framework takes inspirations from key components used in evaluation rubrics in pedagogy. It consists of criteria proposal, pairwise rating by aspect and aggregation of aspect-wise scores. This framework enhances the transparency, accountability and interpretability of the black-box evaluation process.}
    \label{fig:crown_jewel}

\end{figure}

With the use of LLMs as evaluators, a critical question emerges regarding \textit{the extent to which different LLMs can be trusted for reliable evaluation}. To address this question, some recent works focus on the development of meta-evaluation benchmarks \cite{wang2023large, zeng2023evaluating, zheng2024judging}. In these tasks, the basic setting of meta-evaluation involves prompting LLMs one time to ask for a preference among the responses and calculating the agreement with humans. However, this method may not fully reflect LLMs' capability in terms of evaluation: the final output label may be aligned with human preference by chance but with potentially incorrect reasoning. Although interpretable methods such as Chain-of-Thought (CoT; \citealp{wei2023chainofthought}) 
prompting have been adopted in some work to elicit models' explanations, these techniques do not allow a systematic meta-evaluation, due to the uncontrolled reasoning paths adopted for each instance. Moreover, previous work \citep{zeng2023evaluating} has empirically shown that CoT does not bring about consistent performance improvement with step-by-step reasoning, despite offering greater interpretability.


Towards the goals of effectiveness and interpretability, we propose the \textit{DnA-Eval} framework (Figure \ref{fig:crown_jewel}), which is inspired by the use of evaluation rubrics used in pedagogy \citep{dickinson2017values} and the idea of decomposing difficult problems to simpler components for simplifying problem solving \citep{zhou2023leasttomostpromptingenablescomplex}. The framework consists of two main stages of decomposition and aggregation. In decomposition, an LLM either takes the criteria given in instruction as aspects or proposes different aspects when such information is not provided. The LLM performs pairwise scoring for different generations for every aspect. In aggregation, the LLM will be dynamically prompted to propose weightings for different aspects based on their importance in the given instance's context. An external calculation module is executed to compute the weighted sum of scores for different aspects as the overall score and compare the overall scores for two generations to produce a final evaluation judgment.

With our \textit{DnA-Eval} framework, we make the following contributions:
\begin{itemize}
\item We empirically show that our framework leads to \textbf{consistent performance improvement} across different datasets compared to other zero-shot methods such as direct scoring and CoT prompting. Without the need for collecting additional data and conducting any finetuning, it serves as an effective evaluation protocol that works for both proprietary and open-sourced LLMs. 
\item We analyse LLMs' intermediate outputs (model-generated aspects and weightings) to better understand LLMs' capabilities when using them as evaluators. As such, we induce \textbf{greater interpretability} to the LLMs' black-box evaluation process. This results in a better understanding of different LLMs' reliability in evaluating texts.
\item Our framework which is grounded on pedagogical practices, introduces a systematic, modularized reasoning procedure for using LLMs for evaluation. With modularization of stages involved in an evaluation process, we enhance the evaluation process with external calculation during the aggregation stage, shedding light on the design of \textbf{tool-augmented LLM evaluators}. 
\end{itemize}

\section{Related Work}
\paragraph{Automatic Text Evaluation.} The high cost of human evaluation for machine-generated texts has motivated research in developing automatic text evaluation methods. For Natural Language Generation tasks, metrics such as BLEU \citep{papineni2002bleu} and ROUGE \citep{lin2004rouge} scores were used as the dominant approach to evaluate machine-generated text using lexicon overlap based on a candidate reference. Recently, methods like BERTScore \citep{zhang2019bertscore} and BARTScore \citep{yuan2021bartscore} better account for meaning-preserving lexical and compositional diversity and capture semantic information, compared against previous methods which only rely on lexical components. These reference-based methods have limitations in capturing the diversity and richness of human language, especially for subjective open-ended long-form questions \citep{krishna2021hurdles}. 

As such, researchers propose reference-free evaluation methods like iBLEU \citep{sun2012joint} and ParaScore \citep{shen2022evaluation}. 
GPTScore \citep{fu2023gptscore} also leverages the increasing pretrained knowledge and high zero-shot capability of language models. There is ongoing research exploring LLMs as evaluators under reference-free contexts.

\paragraph{LLM-based Text Evaluation.} With the emergence of many powerful LLMs like ChatGPT and GPT-4, increasing work has explored their performance in evaluating generated texts for translation, story generation, paraphrase generation and so on 
\citep{chiang2023can, kocmi2023large, chen2023exploring, wang2023chatgpt, hada2024metal}. These empirical explorations demonstrate the stable performance of LLMs in evaluating a wide range of NLG tasks with different task requirements. 

However, LLMs have limitations and biases during text evaluation, which include \textit{position bias}, where they tend to prefer some positions over others \citep{wang2023large}; \textit{verbosity bias}, where they favor longer responses \citep{zheng2024judging}; \textit{self-enhancement bias}, where they favor or disfavor self-generated answers \citep{zheng2024judging}; and \textit{style bias}, where they value style of texts generated more than content\citep{wu2023style}. 

In light of these limitations, researchers are exploring ways to improve LLMs' evaluation capability. Previous work like G-Eval prompts LLMs to generate chain of thoughts for evaluation steps and take the weighted sum over probabilities for different scores \citep{liu2023gpteval}. Kim et al. proposes Prometheus (\citeyear{kim2023prometheus}) and Prometheus~2 (\citeyear{kim2024prometheus2opensource}), evaluation-specific open-source models with fine-tuning on the feedback to effectively induce fine-grained evaluation capability. More recent approaches like Chain-of-Aspects \citep{gong2023coascore} and Branch--Solve--Merge \citep{saha2023branchsolvemerge} offer new paradigms for LLMs to decompose multi-faceted language evaluation tasks. In addition, the ChatEval framework \citep{chan2023chateval} is proposed to increase LLMs' evaluation capability through multi-agent debate.

On top of these methods, our work proposes a generalizable evaluation framework with careful design for both decomposition and aggregation stages. Under this framework, LLMs' evaluation performance is \textit{consistently} improved on both proprietary and open-sourced models while providing higher interpretability at the same time. Compared to previous methods, we achieve better evaluation performance and greater interpretability without loading multiple models \citep{chan2023chateval}, collecting more data \citep{kim2023prometheus}, or conducting  finetuning \citep{kim2024prometheus2opensource}.

\paragraph{Meta-Evaluation of LLMs as Evaluators.} As a newly emerging research area, there are only a few benchmarks for meta-evaluation of LLMs as evaluators. Therefore, how reliable LLMs are as evaluators still remains an important research question worth investigating.

To build meta-evaluation benchmarks, recent work leverages on previous meta-evaluation datasets \citep{fu2023gptscore}, carries out small-scale expert annotation for specific tasks \citep{wang2023large} and crowd-sources human annotation \citep{zheng2024judging}. Meta-evaluation methods include computing correlations with human ratings \citep{gong2023coascore}, calculation of agreement with human labels \cite{wang2023large, zheng2024judging, zeng2023evaluating}, and performing meta-evaluation using agent debate \citep{chern2024can}. However, few works focus on the interpretability of the meta-evaluation process: high agreement or correlation of the final judgment with human labels does not necessarily mean a strong evaluation capability, as the intermediate reasoning process may be flawed. This is especially true where there are only two possible answers for preference agreement computation. The LLM may make the aligned preference with human by chance with incorrect reasoning. 

Although there exists some previous work adopting CoT prompting in their experiments to provide more interpretability to the black-box evaluation process, these have been shown to be ineffective to improving the general performance of LLMs' evaluation capability \citep{zeng2023evaluating}. Our work is able to achieve \textit{performance improvement} while \textit{enhancing interpretability}.

\section{DnA-Eval Framework}
\begin{figure*}\includegraphics[width=\textwidth]{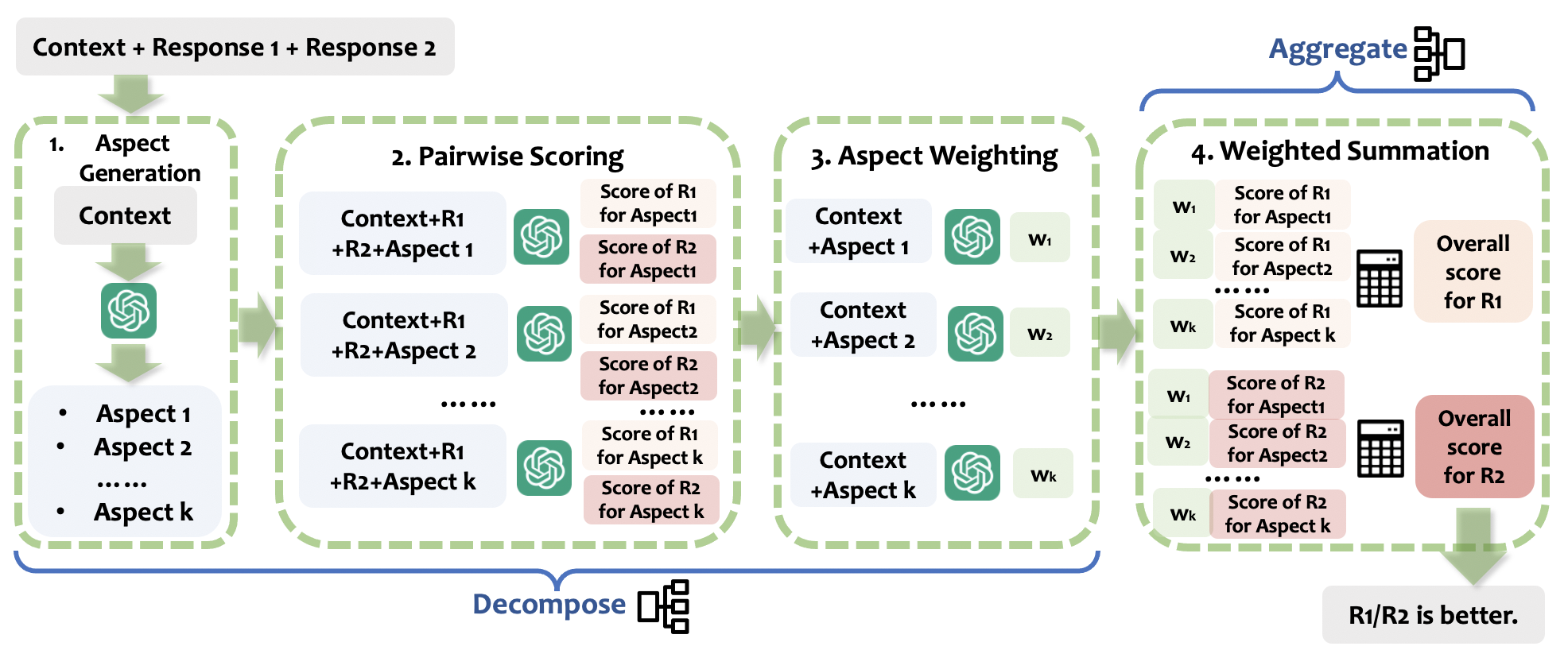}
  \caption{\small Different stages of \textit{DnA-Eval}. In the decomposition stage, LLMs are provided with the context to propose $k$ different evaluation aspects. These aspects are combined with the context and candidate responses for LLMs to generate pairwise scores for each aspect. LLMs will also be prompted to provide respective weightings for each aspect with the given context. In the aggregation stage, external computing tool can be used to calculate the overall scores for each response and make comparison to decide on the better response.} \vspace{-0.8em}
  \label{fig:pipeline}
\end{figure*}
The benefits of scoring rubrics in evaluation processes have been noted in previous research, which are facilitated learning, increased consistency and more valid evaluation of complex competencies \citep{jonsson2007use}. Inspired by its extensive applications in pedagogy, we establish the \textit{DnA-Eval} framework for using LLMs as evaluators. The framework bases on core elements in scoring rubrics, which are the decomposition and aggregation stages (Figure~\ref{fig:pipeline}).

\subsection{Aspect Generation}
Appropriate criteria is the key to effective evaluation rubrics \citep{brookhart2018appropriate}. They serve as clear guidelines for aspects to be evaluated and provide greater transparency in how a final evaluation judgment is derived. The criteria aspects are determined by specific requirements of different instances for different tasks. The set of evaluation aspects for the $i$-th instance can be formulated as:
\begin{equation}
    \label{eq:eval}
    \mathbf{A}_i=\{A_{i1}, A_{i2}, \ldots, A_{ik}\}, \text{  for } i=1,2,\ldots,n
\end{equation}
where $A_{ij}$ denotes the $j$-th evaluation aspect for the $i$-th instance, $k$ is the total number of aspects, and $n$ is the total number of instances.

In previous evaluation tasks, there are two possible scenarios where the evaluation aspects can be \textit{predefined} or \textit{unspecified}. In the first scenario, there exists an explicitly-defined set of criteria for the evaluation task; i.e., each and every instance in the given dataset will be evaluated using the same aspect set. In the second setting, there are no clearly defined aspects provided. Under such cases, we propose dynamic aspect generation, whereby an LLM is prompted to generate the values of $\mathbf{A}_i$ given the problem context of the $i$-th instance and a predetermined number of aspects, denoted by $k$.

\subsection{Pairwise Scoring by Aspect}


There are two general frameworks for using LLM-as-a-judge in existing work. The first one is pairwise comparison where LLMs are prompted to determine if the first or the second response is better given a query \citep{zheng2024judging}. The second framework is evaluation by scoring where LLMs are tasked to provide numerical scores for different responses. The final decision about the better response is made by comparing the scores generated by LLMs \citep{wang2023large}. Taking the respective pros and cons of these two frameworks into consideration, we adopt the approach of \textit{pairwise scoring} in our framework. This combines the strengths of both methods --- namely, the ability to capture subtle differences in pairwise comparison framework \citep{liu2024aligning}, and the higher scalability in evaluating multiple candidates and higher interpretability in the single answer scoring framework \citep{zheng2024judging}. We formulate our pairwise scoring mechanism as follows:
\begin{align*}
    \label{eq:scores}
    &\mathbf{S}_i=\{\mathbf{S}^{(1)}_i, \mathbf{S}^{(2)}_i\}, \text{  for } i=1,2,\ldots,n \\
    &\mathbf{S}^{(1)}_i=\{S^{(1)}_{i1}, S^{(1)}_{i2}, \ldots, S^{(1)}_{ik}\} \\
    &\mathbf{S}^{(2)}_i =\{S^{(2)}_{i1}, S^{(2)}_{i2}, \ldots, S^{(2)}_{ik}\}
    \tag{2}
\end{align*}

\noindent where $\mathbf{S}_i$ is the generated scores for different responses for the $i$-th instance in the dataset along different aspects, consisting of two score sets ($\mathbf{S}^{(1)}_i$ and $\mathbf{S}^{(2)}_i$) for the response candidates. $\mathbf{S}_i$ may include more than two score sets when the evaluation is conducted for more than two candidates. $S^{(m)}_{ij}$ denotes the score value for the $m$-th candidate of the $i$-th instance along the $j$-th aspect.

\subsection{Aggregation}
For each instance, the score set $\mathbf{S}_i$ with $k$ pairs of scores for  the $k$ different aspects will be generated in the decomposition stage. Previous work \citep{gong2023coascore} passes aspect-wise score pairs as contexts in prompts for LLMs to provide the overall scores. However, it has been shown that LLMs may struggle to solve computation problems \citep{zhang2024evaluating}. Therefore, we augment the framework with an external calculation module. We define an aggregation function $f$ to compute the final score for each response. The aggregation will take the weighted sum of scores for each aspect:
\begin{equation}
    \label{eq:objective}
    f(\mathbf{S}^{(m)}_i)=\sum_{j=1}^k w_{ij}S^{(m)}_{ij}
    \tag{3}
\end{equation}
where $w_{ij}$ is the weightage for the $j$-th aspect of the $i$-th instance. It can be obtained by prompting the LLMs for a percentage weightage indicating the importance for a specific aspect and instance. 

After aggregating aspect-wise scores to the overall scores, the predicted label for the $i$-th instance is determined by comparing the overall scores:
\begin{equation}
 \tilde{y}_i = \begin{cases} 
 1 & f(\mathbf{S}^{(1)}_i)>f(\mathbf{S}^{(2)}_i) \\  
 2 & f(\mathbf{S}^{(1)}_i)<f(\mathbf{S}^{(2)}_i) \\  
 0 & f(\mathbf{S}^{(1)}_i)=f(\mathbf{S}^{(2)}_i) 
 \end{cases} 
 \tag{4}
\end{equation}
where $1$ indicates Response~1 is better, $2$ indicates Response~2 is better and $0$ indicates a tie.

\section{Experiments}
We conduct the experiments on four different meta-evaluation benchmarks. We select more recent meta-evaluation benchmarks (published in 2023 or later) to mitigate the data leakage problem \citep{jiang2024investigating}. In these benchmarks, each instance is annotated with a human preference label indicating which of the two responses is better. The four benchmarks cover two possible scenarios where a fixed set of criteria is given or not provided to human annotators in the evaluation process as summarized in Table~\ref{tab:datasets}. They cover a wide variety of task categories, including writing, math, knowledge, common sense, coding and summarization.
\begin{table}[!ht]
\def\arraystretch{1.6}
\resizebox{\columnwidth}{!}{
\begin{tabular}{ccc}
\hline
\textbf{Dataset}  & \textbf{Defined Criteria} & \textbf{Presence of Ties} \\ \hline
\textbf{FairEval} &       \yeslogo                    &          \yeslogo                 \\ \hline
\textbf{MT-Bench} &       \yeslogo                    &          \yeslogo                 \\ \hline
\textbf{LLMBar}   &     \nologo                      &         \nologo                  \\ \hline
\textbf{Instrusum}  &     \nologo                      &        \yeslogo                   \\ \hline
\end{tabular}}
\caption{\small Summary of key features of meta-evaluation datasets used in our experiments. FairEval and MT-Bench have predefined criteria aspects while LLMBar and Instrusum do not provide such aspects to human annotators when collecting preferences. There are tie cases in FairEval, MT-Bench and Instrusum datasets but there are no tie cases in LLMBar.}
\label{tab:datasets}
\end{table}
\paragraph{FairEval} \citep{wang2023large} holds a collection of 80 questions with two responses from Vicuna-13b and ChatGPT for each question. Annotators were asked to label which response is better or if it is a tie given four perspectives: helpfulness, relevance, accuracy and level of details. 

\paragraph{MT-Bench} \citep{zheng2024judging} contains 80 questions with responses from 6 different models (GPT-4, GPT-3.5, Claude-v1, Vicuna-13B,  Alpaca-13B and LLaMA-13B). They are labelled with preference by graduate students along the six dimensions of helpfulness, relevance, accuracy, creativity, depth and detail. As it is computationally expensive to run inference over the entire dataset, due to budget constraint, we perform stratified random sampling for 400 single-turn samples, covering of all unique questions in the dataset.

\paragraph{LLMBar} \citep{zeng2023evaluating} consists of 419 questions that can be objectively evaluated for the instruction following ability. We take the adversarial set of 319 instances in LLMBar benchmark for our experiment. The adversarial set holds adversarially crafted instances which are more prone to confuse less adept evaluators. Different LLMs have remarkable difference in evaluation capability on the challenging adversarial set.


\paragraph{InstruSum} \citep{liu2023benchmarking} comprises 100 human-written articles and summary requirements. Each article is accompanied with LLM-generated or hybrid LLM-human summaries annotated with human ratings on the overall quality. There are five systems evaluated in InstruSum and we select summaries from \texttt{GPT-3.5-turbo-0301} and \texttt{GPT-4-0314} to construct pairs used for our experiments as these two systems have similar text generation capability among the five system options.

\subsection{Experimental Setup}
\begin{table*}[t]
\centering
\def\arraystretch{1.8}
\resizebox{\textwidth}{!}{  
\begin{tabular}{ccccccccc}
\hline
\textbf{}                       & \multicolumn{4}{c}{\textbf{LLMBar-Adversarial}}                                   & \multicolumn{4}{c}{\textbf{InstruSum}}                                         \\ \hline
                                & ChatGPT            & GPT-4              & LLaMa2-13B             & Mistral-7B            & ChatGPT               & GPT-4              & LLaMa2-13B          & Mistral-7B            \\ \hline
Direct Scoring                  & $29.8$               & $70.8$               & $29.8$               & $32.9$               & $49.0\:52.2$                  & $38.0\:40.0$               & $53.0\:58.9$            & $17.0\:11.1$               \\ \hline
Scoring with CoT                   & $24.8$                 & $75.2$                & $33.9$                & $\mathbf{43.3}$                & $23.0\:21.1$           & $48.0\:52.2$               & $47.0\:52.2$              & $\mathbf{30.0}\:\mathbf{33.3}$                \\ \hline
DnA-Eval (ours) & $\mathbf{33.5}$       & $\mathbf{77.1}$       & $\mathbf{34.2}$       & $39.2$       & $\mathbf{60.0}\:\mathbf{64.4}$                   & $\mathbf{53.0}\:\mathbf{57.8}$       & $\mathbf{60.0}\:\mathbf{66.7}$   & $25.0\:21.1$       \\ \hline
Chain of Aspects (Ablation)     & $30.4$\reduce{}          & $75.5$\reduce{}            & $33.9$\reduce{}            & $27.6$\reduce{}            & $43.0$\reduce{} $44.4$\reduce{}               & $51.0$\reduce{} $52.2$\reduce{}            & $48.0$\reduce{} $53.3$\reduce{}   & $11.0$\reduce{} $3.3$\reduce{}            \\ \hline
                                & \multicolumn{4}{c}{\textbf{MTBench400}}                                           & \multicolumn{4}{c}{\textbf{FairEval}}                                             \\ \hline
                                & ChatGPT            & GPT-4              & LLaMa2-13B             & Mistral-7B            & ChatGPT               & GPT-4              & LLaMa2-13B          & Mistral-7B            \\ \hline
Direct Scoring                  & $58.0\:71.8$          & $\mathbf{67.8}\:74.4$          & $53.8\:71.4$          & $53.8\:61.1$          & $53.8\:60.6$            & $46.3\:42.4$          & $\mathbf{46.3}\:\mathbf{56.1}$       & $\mathbf{52.5}\:\mathbf{62.1}$ \\ \hline
Scoring with CoT                   & $58.0\:71.4$          & $61.3\:76.7$          & $54.3\:72.1$          & $\mathbf{58.0}\:66.7$ & $42.5\:36.4$            & $50.0\:54.5$          & $43.8\:53.0$       & $48.8\:54.5$          \\ \hline
DnA-Eval (ours) & $\mathbf{59.8}\:\mathbf{74.8}$  & $65.3\:\mathbf{78.4}$  & $\mathbf{56.3}\:\mathbf{74.8}$  & $56.8\:\mathbf{67.4}$  & $\mathbf{56.3} \: \mathbf{65.2}$             & $\mathbf{51.3}\:\mathbf{59.1}$  & $\mathbf{46.3}\:\mathbf{56.1}$      & $\mathbf{52.5}\:\mathbf{62.1}$ \\ \hline
Chain of Aspects (Ablation)    & $59.8\:73.1$\reduce{}       & $66.8\:76.7$\reduce{}     & $55.0$\reduce{} $73.1$\reduce{}    & $53.3$\reduce{} $61.1$\reduce{}    & $51.3$\reduce{} $56.1$\reduce{} & $48.8$\reduce{} $56.1$\reduce{}    & $41.3$\reduce{} $50.0$\reduce{} & $50.0$\reduce{} $60.6$\reduce{}    \\ \hline
\end{tabular}}
\caption{\small Percentage agreement with human preference label of each LLM on different meta-evaluation benchmarks. For InstruSum, MTBench and FairEval, we report the agreement with (first number) and without (second number) tie cases in each cell. \reduce{} marks the situation where the ablation setting (replacing weighted sum aggregation with prompted aggregation) leads to a drop in performance, suggesting LLMs' limitations in aggregating the scores during the evaluation process.} 
\label{tab:result}
\end{table*}

\paragraph{Models.} We select two proprietary LLMs (GPT-3.5 and GPT-4) and two open-sourced LLMs (Llama2-13B and Mistral-7B-Instruct-v0.2) for a comprehensive exploration. This also allows meaningful comparisons of the evaluation capability between these two general classes. We select the \texttt{06-13} model version for GPT-3.5 and GPT-4 to mitigate the data leakage issue (see Appendix \ref{sec:leakage}).

\paragraph{Baselines.} We compare the performance of our proposed framework to two zero-shot baselines for a fair comparison. One baseline is the \textit{direct scoring method} which asks the models for the overall score for each response directly. The second baseline is the \textit{Chain-of-Thought (CoT) method} which asks models to provide explanations first followed by the overall score for each of the two responses.

\paragraph{Prompts.} We adopt the same prompting templates from the original experiment of each benchmark in the direct scoring method as they are carefully designed for the specific requirements of each task. For the CoT method and aspect generation, we follow the prompting templates in the work of \citet{zeng2023evaluating} by asking for explanations before scores and asking for three relevant questions in evaluating the instance.

For aspect generation and aspect weighting stages, we include only the question context of the instance but not the responses in the prompts. This is because in real world situations, the design of the evaluation rubrics is usually task-specific without the need of knowing the responses to the question. For aspect-wise scoring, we ask for scores of different aspects in separate inferences since LLMs may be subject to anchoring effects for multi-attribute evaluation \citep{stureborg2024large}, where the generated scores in the same inference are correlated with one another.

\subsection{Results}
From Table \ref{tab:result}, we observe that \textit{DnA-Eval} generally outperforms both baselines of direct scoring and CoT for both proprietary and open-source models across different datasets. The performance reaches39.6\% (agreement of 42.4\% with direct scoring and 59.1\% with \textit{DnA-Eval} for GPT-4 on FairEval benchmark). Our results also corroborate the findings from previous work \citep{zeng2023evaluating} that CoT method does not bring about consistent improvement to LLMs' evaluation capability and sometimes even worsens it. This shows that our framework is better in terms of being both \textit{interpretable} and \textit{effective} at the same time. 


\subsection{Ablation Study} 
We also conduct an ablation study to investigate the effectiveness of the weighted sum approach in aggregation stage. In the ablation experiments, we pass the pairwise scoring for each aspect to LLMs as part of the prompt and ask the models to generate an overall score for each response respectively. It is a common practice to aggregate aspect-wise scores in previous work \citep{gong2023coascore, saha2023branchsolvemerge} involving multi-aspect evaluation.

For almost all LLMs and benchmarks tested, using an external calculator to compute the weighted sum achieves a higher agreement with human than directly passing aspect-wise scores as prompts to LLMs. This suggests \textit{the limitation of LLMs in mathematical aggregation} during the evaluation process. The results also show that our method using LLM-proposed weightings and an external computation module helps to address such limitation.

\paragraph{Qualitative Analysis.} To better understand where the performance improvement comes from, we conduct qualitative analysis for cases where direct scoring gives an incorrect evaluation, but where \textit{DnA-Eval} provides a correct evaluation. We identify two main categories of improvement among these instances, which are \textit{(i) more accurate prioritization of different aspects} and \textit{(ii) more subtle judgment} (Appendix \ref{app:qual}). 
\begin{figure*}\includegraphics[width=\textwidth]{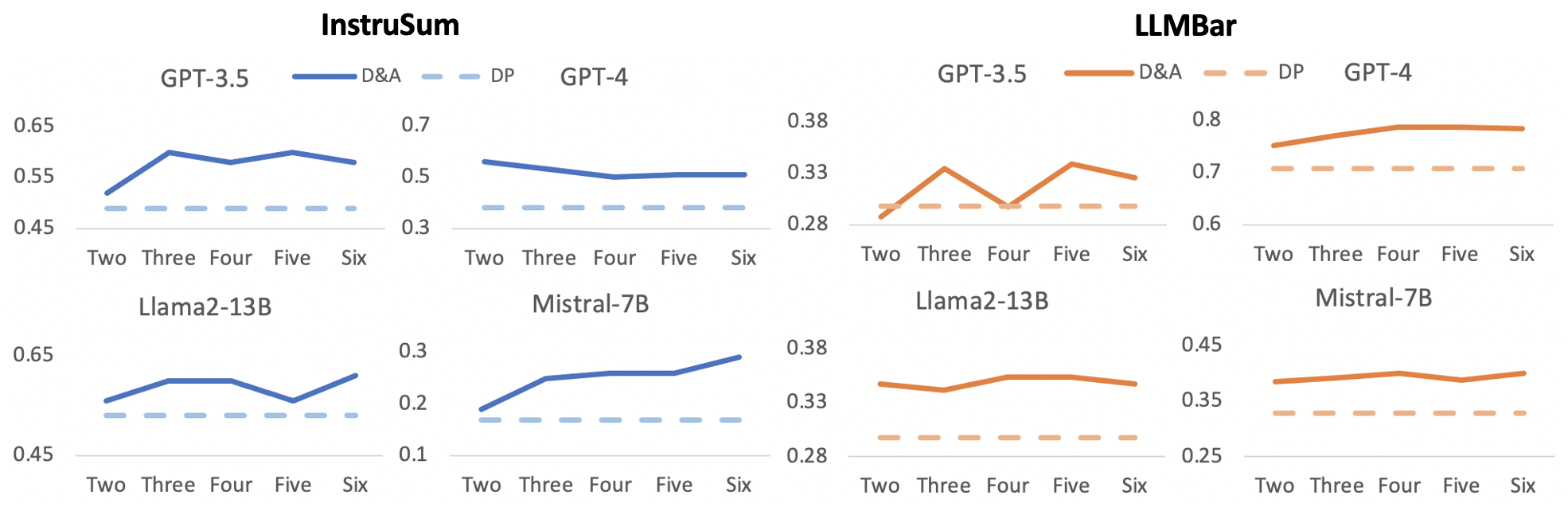}
  \caption{\small Agreement with human annotators with varied number of aspects. We also report the baseline performance of direct prompting in dashed lines. Our framework generally outperforms the baseline regardless of number of aspects chosen.} 
  \label{fig:numberofaspects}
\end{figure*}

\section{Analyses}
\textit{DnA-Eval} produces intermediate outputs like LLMs' self-generated aspects and weightings for different aspects. This offers practitioners an opportunity to interpret and evaluate the intermediate steps of LLMs' evaluation process. Therefore, we perform further analyses for self-generated aspects and weightings from different language models.

\subsection{Model-Generated Aspects}
\paragraph{Effect of Number of Aspects.} We vary the number of aspects generated by models to study the effect of the \textit{quantity} of aspects during the decomposition stage. Our results (Figure \ref{fig:numberofaspects}) shows that D\&A generally outperforms the baseline across a range of aspect numbers, demonstrating the effectiveness of our method regardless of the number of aspects chosen. Most of dataset--model combinations show an upward trend with fluctuations, suggesting better evaluation performance with increased number of aspects. However, a higher number of aspects does not always lead to greater evaluation performance (e.g. decrease in performance of GPT-4 on Instrusum). This indicates that the optimal number of aspects \textit{depends on the specific task and the LLM evaluator used}. Therefore, practitioners are suggested to conduct some experiments on a pilot dataset to study the most cost-effective choice for the number of aspects.

\paragraph{Quality of Aspects.} 
We recruit crowdworkers on Amazon Mechanical Turk to evaluate LLM-generated aspects. We ask three crowdworkers to rate the \textit{relevance}, \textit{clarity} and \textit{comprehensiveness} of model-generated aspects independently for each instance. The rating is on a Likert scale of 1 to 5. We randomly sample 50 instances from each dataset and report the average scores along each dimension for different models. 

All four models in our experiments achieve an above-average performance with scores higher than 4 for all three dimensions, suggesting \textit{LLMs are capable of generating evaluation aspects of good quality}. However, there exist some nuanced differences across different models (Table \ref{tab:human_eval1}). Proprietary models like ChatGPT and GPT-4 generally generates aspects that are more relevant, clearer and more comprehensive than open-sourced models. ChatGPT performs the best for relevance (4.95) and clarity (4.93) and GPT-4 outperforms other models in terms of comprehensiveness (4.84). On the other hand, Llama2-13B model performs the worst in generating evaluation aspects.

We also collect free-text explanations from crowdworkers to better understand their evaluation for model-generated aspects. Annotators identify different levels of relevance for different aspects generated by the models. They consider certain aspects as \textit{more crucial} while others being \textit{important but less relevant}. The varying relevance of aspects identified by human annotators justifies the aspect weighting stage in our framework. Moreover, for aspects with relatively lower ratings for clarity, annotators comment that there is a need for more \textit{specific guidance in terms of examples or illustrations} (e.g. what constitutes `a balanced view' mentioned in one criterion aspect). Additionally, they suggest there could be further breakdown of generated aspects to sub-aspects. In explanations for aspect weightings, annotators also justify the reasons for why some aspects are more \textit{crucial} than others, indicating the varying importance of model-generated aspects.

\begin{table}[!ht]
\def\arraystretch{1.4}
\resizebox{\columnwidth}{!}{
\begin{tabular}{@{}cccc@{}}
\toprule
\textbf{Model} & \textbf{Relevance} & \textbf{Clarity} & \textbf{Comprehensiveness} \\ \midrule
ChatGPT        & \textbf{4.95}      & \textbf{4.93}    & 4.80                       \\ \midrule
GPT-4          & 4.89               & 4.90             & \textbf{4.84}              \\ \midrule
Llama2-13B         & 4.70               & 4.78             & 4.64                       \\ \midrule
Mistral-7B        & 4.89               & 4.87             & 4.71                       \\ \bottomrule
\end{tabular}}
\small\caption{\small Average human ratings for aspects generated by ChatGPT, GPT-4, Llama2-13B and Mistral-7B along the dimensions of relevance, clarity and comprehensiveness.} \vspace{-0.8em}
\label{tab:human_eval1}
\end{table}

\subsection{Model-Generated Weights}
\begin{figure}[!ht]
    \centering \includegraphics[width=\linewidth]{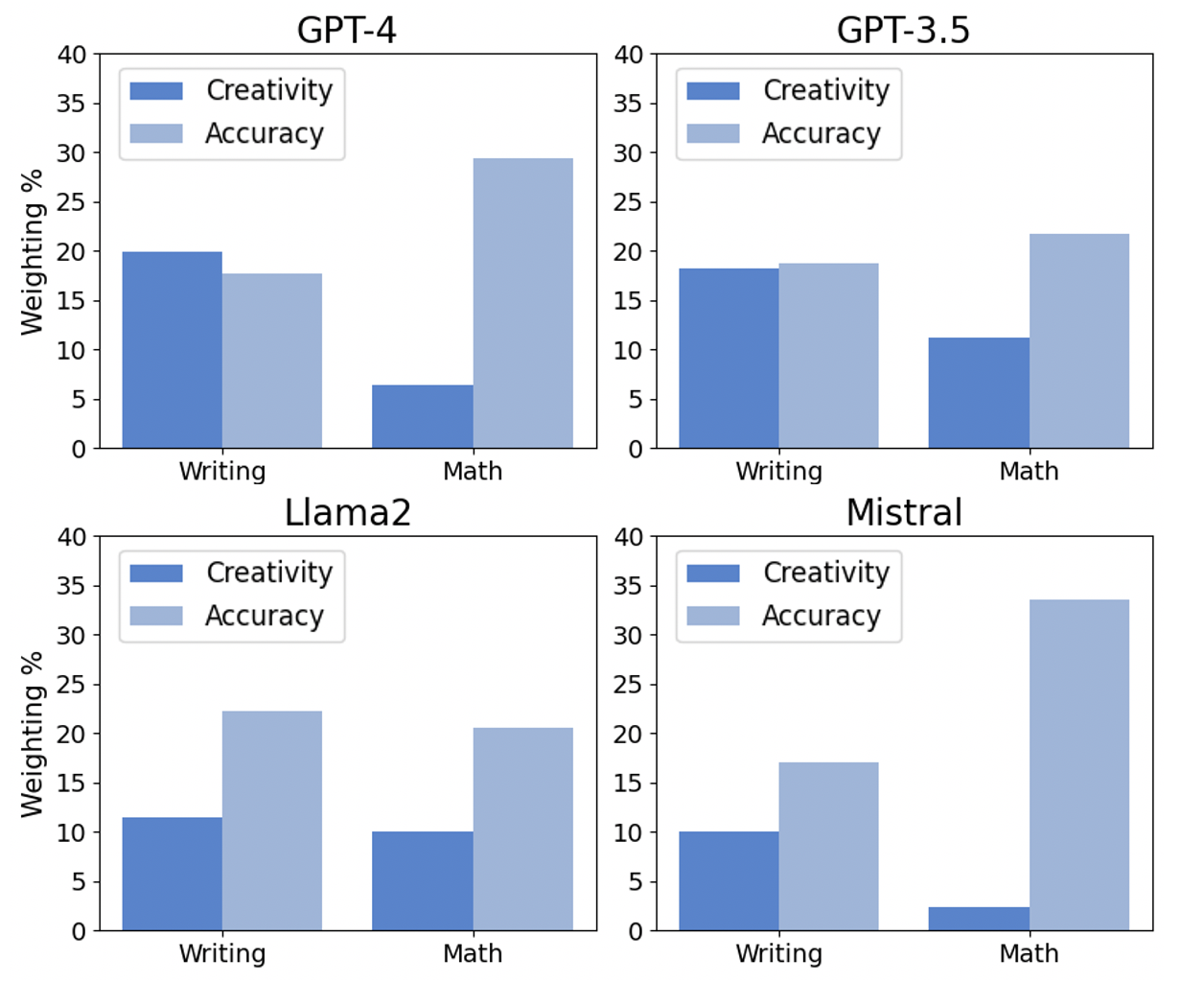}
    \caption{\small Average model-generated weightings for writing and math tasks in MTBench dataset. We report weightings for creativity and accuracy which are task-dependent dimensions. 
    The figure shows all models are able to assign lower weightings for creativity and higher weightings for accuracy for math problems compared to writing tasks. This suggests about their capability in generating weightings that are helpful for evaluation.}\vspace{-0.8em}
    \label{fig:task}
\end{figure}

\paragraph{Weights for Different Tasks.} 
To better understand model-generated weights, we leverage on the annotated task categories in MTBench dataset and analyze the average weights assigned for each aspect for different task categories. Our findings suggest that in general, relevance, accuracy and helpfulness are assigned higher weights by all models for all tasks (Appendix \ref{app:task}). Also, the importance of some aspects is task-dependent (e.g. creativity for writing tasks) and LLMs are able to adjust their weightings for different tasks and prioritize the more important aspects (Figure \ref{fig:task}).
\paragraph{Agreement with Human.}

For evaluation of model-generated weightings, it is difficult to define ratings on a Likert scale and ask human evaluators to numerically rate the quality of different weightings. Therefore, we convert the weightings from models and humans to ranks and then compute the top-k Kendall's~$\tau$ ranking distance \citep{fagin2003comparing} between models' and crowdworkers' rankings. A lower distance indicates a higher weighting similarity. For comparison, we also compute the Kendall's $\tau$ between two different human annotators.

\begin{figure}[!ht]
    \centering \includegraphics[width=\linewidth]{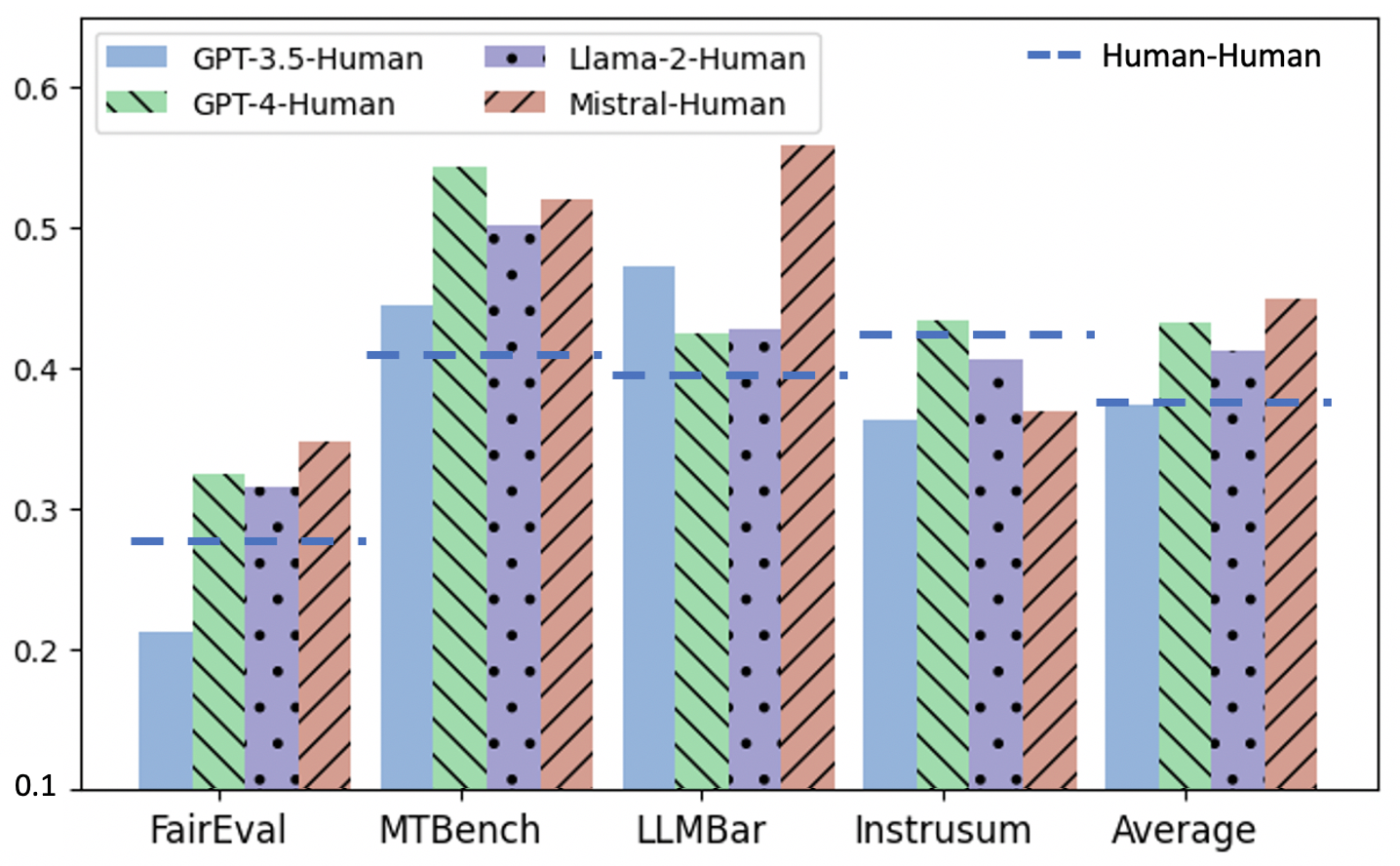}
    \caption{\small Kendall's~$\tau$ distance for aspect weightings between different language models and human. We visualize the rank distance between two different human annotators in dotted lines for a comparison.}\vspace{-0.8em} 
    \label{fig:weighting}
\end{figure}

In general, we see that there still exists more divergence between LLM's weightings and human's weightings (Figure \ref{fig:weighting}) with higher Kendall's~$\tau$ distances between LLM and human than that between humans (e.g. on MTBench and LLMBar) with a few exceptions where LLMs' weightings are more aligned with human's weightings (ChatGPT on FairEval; ChatGPT, Llama2 and Mistral on InstruSum). On average, ChatGPT's weightings are most aligned with human's and Mistral gives the most different weightings from human. 


\section{Conclusions}
We propose the \textit{DnA-Eval}, an effective and interpretable framework to use LLMs as evaluators. From our experiments on meta-evaluation datasets with various features (e.g. presence of tie cases, presence of user-defined criteria) across different domains (e.g. writing, coding, summarization), we illustrate the effectiveness of the framework in enhancing LLMs' evaluation capability. We combine natural language reasoning (decomposition stage) with formalized symbolic reasoning (aggregation stage) in our proposed framework to introduce higher flexibility, reliability and verifiability. Moreover, our analyses provide interpretable insights on different LLMs' evaluation capability in terms of aspect generation and aspect weighting. Such module-level analyses are able to shed light on  multi-agent or human--LLM collaboration in evaluating texts.


\section*{Limitations}
From our experimental results, we show that \textit{DnA-Eval} not only provides higher interpretability of LLMs' evaluation process but also leads to performance improvement compared with direct scoring method. However, there are additional costs incurred with longer input lengths and increased number of inferences for aspect-wise evaluation and weighting generations (Appendix \ref{app:cost}).

Moreover, we set a fixed number of aspects (three aspects) in our experiments. The number of aspects that are relevant may be context-dependent and may vary from case to case. Therefore, future work could explore what is the optimal number of aspects and investigate the effectiveness of dynamic aspect generation with unspecified number of aspects.

In addition, we evaluate the performance of baselines and our method using agreement with human preference labels. This is the most common approach adopted in current meta-evaluation work. However, human preference labels may not be the gold label all the time and agreement with human preference may not be the most accurate way to measure LLMs' evaluation capability. For example, in some cases labeled as `ties' by human, LLMs are able to identify nuanced differences and pick the slightly better answer, demonstrating super-human level evaluative capability. We  observed this during experimentation with GPT-4 on MTBench dataset where the model identified subtle differences in two responses unnoticed by humans. Therefore, there is no improvement with tie cases included but there exists improvement with tie cases excluded when applying our framework.

\section*{Ethics Statement}
This study has been approved by the Institutional Review Board (IRB) at the researchers' institution, and we obtained participant consent with a standard institutional consent form. One potential ethical concern of using LLMs as evaluators is the stereotypes and biases existing in LLMs such as political bias, gender bias, cultural bias and so on. Although our work mainly serves as a new framework to improve LLMs' evaluation capability with greater interpretability, we still acknowledge these potential ethical concerns that may come with using LLMs as judges.

\section*{Acknowledgements}
We are thankful to Do Xuan Long, Tongyao Zhu as well as anonymous reviewers for their helpful feedback. Minzhi Li is supported by the A*STAR Computing and Information Science (ACIS) Scholarship. 
This research is also supported by the National Research Foundation, Singapore under its AI Singapore Programme (AISG Award No: AISG2-GC-2022-005); and NUS-NCS Joint Laboratory (A-0008542-00-00). Any opinions, findings and conclusions or recommendations expressed in this material are those of the author(s) and do not reflect the views of National Research Foundation, Singapore.
	
\bibliography{custom}

\appendix
\label{sec:appendix}
\section{LLM Inference Setting}
\label{sec:expsetting}
\paragraph{Temperature Setting.} We set temperature to 0 for classification tasks to ensure reproducibility.

\paragraph{Prompts.} 
\begin{itemize}
    \item \textbf{Direct Scoring:} We adopt the same prompting templates from the original experiment of each benchmark for the direct scoring method as they are carefully designed for the specific requirement of each task. The prompts contain the instance context, candidate responses and evaluation instruction.
    \item \textbf{CoT Prompting:} We ask the models to provide an explanation and then an overall socre for each of the response candidates in the instruction. The prompts contain the instance context, candidate responses and evaluation instruction.
    \item \textbf{Aspect Generation:} When criteria aspects are not given in the cases of LLMBar and InstruSum, we follow the prompting templates for metrics generation strategy in the work of \citet{zeng2023evaluating} by asking the models to \textit{propose three concise questions about whether a potential output is a good output for a given instruction}. The prompts contain the instance context and aspect generation instruction.
    \item \textbf{Aspect-wise Scoring:} The prompting templates we use are similar to direct scoring. The only difference is that we pass the pre-defined criteria or the model-generated aspect to the model for pairwise scoring on top of the instance context, candidate responses and evaluation instruction.
    \item \textbf{Weighting Proposal:} We formulate our instruction as \textit{``Please propose respective importance weightage for three aspects in evaluating the summary.''} The prompts contain the instance context and model-generated aspects. We further specify some requirements for the weighting outputs: \textit{``1) The weightages should be in percentage form and sum up to 100\%; 2) You should directly give the weightages without any other words; 3) You should give weightages in the same line, separated by space.''}
    
\end{itemize}

\section{Robustness Check}
Due to relatively small number of instances in each meta-evaluation dataset, there is little statistical significance in performance difference between the baseline method and our method. Therefore, we repeat the experiments at two other different seeds and calculate the statistical significance. From Table \ref{tab:significance}, we can see that the performance improvement achieved by \textit{DnA-Eval} is generally statistically significant on most of model--dataset combinations. 

\begin{table}[!ht]
\def\arraystretch{1.4}
\resizebox{\columnwidth}{!}{
\begin{tabular}{@{}ccccc@{}}
\toprule
                 & \textbf{FairEval} & \textbf{MTBench} & \textbf{InstruSum} & \textbf{LLMBar} \\ \midrule
\textbf{ChatGPT} & **                & ***              & ***             & Not Significant \\
\textbf{GPT-4}   & **                & **               & ***           & ***             \\
\textbf{Llama2-13B}  & Not Significant   & **               & **             & **              \\
\textbf{Mistral-7B} & Not Significant   & ***              & ***           & ***             \\ \bottomrule
\end{tabular}}
\caption{Significance test results for \textit{DnA-Eval} and Direct Scoring method. * p<0.1, ** p<0.05, *** p<0.01}
\label{tab:significance}
\end{table}

\section{Data Leakage Analysis}
\label{sec:leakage}
There is minimal likelihood of data leakage if the release date of the model is before the release date of the dataset. From Table \ref{tab:dates}, most dataset--model combinations in our experiments are not subject to data leakage. However, there may exist data leakage for testing Llama2-13B and Mistral-7B on the FairEval dataset. Such data leakage issue may be the cause for no improvement of our \textit{DnA-Eval} method compared to direct scoring method. There is also a slight chance of data leakage for testing Mistral-7B on LLMBar and InstruSum.
\begin{table}[!ht]
\def\arraystretch{1.8}
\resizebox{\columnwidth}{!}{
\begin{tabular}{@{}cccc@{}}
\toprule
\textbf{Dataset} & \textbf{Release Date} & \textbf{Model} & \textbf{Release Date} \\ \midrule
FairEval         & May 2023              & GPT-3.5-0613   & Jun 2023              \\ \midrule
MTBench          & Dec 2023              & GPT-4-0613     & Jun 2023              \\ \midrule
LLMBar           & Nov 2023              & Llama2-13B         & Jul 2023              \\ \midrule
InstruSum        & Nov 2023              & Mistral-7B-Instruct-v0.2     & Dec 2023              \\ \bottomrule
\end{tabular}}
\caption{Release dates for different datasets and models experimented.}
\label{tab:dates}
\end{table}

\newpage
\section{Human Evaluation Collection}
\subsection{Qualification}
We recruit crowdworkers on Amazon Mechanical Turk to evaluate the \textit{quality} of LLM-generated aspects and weightings. To ensure data quality, we require the annotators to have an accepted number of tasks higher than 500 and an approval rate higher than 98\%. Crowdworkers who fulfilled these criteria went through a qualification round which contains exactly the same questions in the actual round. Their submissions for the qualification round were manually verified by the authors and qualified workers were given access to the actual round. We pay all annotators a fair wage (US\$15 per hour) above the federal minimum.
\subsection{Human Annotation Instructions}
We provide human annotators with detailed instructions and examples for aspect evaluation and aspect weighting.
\subsection{Instruction for Aspect Evaluation}
You will rate the relevance, clarity, and comprehensiveness of different aspects in evaluating responses to a question.
\begin{enumerate}
  \item \textbf{Relevance:} Are the aspects relevant to the question? Relevant aspects should directly align with the objectives and goals of the evaluation.
  \textbf{Example:}\\
  \textbf{Question:} Solve for x in the equation 3x + 10 = 5(x - 2).\\
  \textbf{Aspect~1:} Answer accuracy \quad \textbf{Relevance:} 5\\
  \textbf{Explanation:} Answer accuracy is very relevant as the primary goal of solving an equation is to find the correct value or values of the variable.\\
  \textbf{Aspect~2:} Level of humor \quad \textbf{Relevance:} 1\\
  \textbf{Explanation:} Level of humor is very irrelevant as because humor has no bearing on the mathematical process involved in solving the equation.\\
  \item \textbf{Clarity:} Are the aspects clearly defined and easily understood by potential evaluators? Clear aspects should have no ambiguity or vagueness.
  
  \textbf{Example:}\\
  \textbf{Question:} Write a poem in Shakespearean style.\\
  \textbf{Aspect~1:} Application of Shakespearean style \quad \textbf{Clarity:} 5\\
  \textbf{Explanation:} Application of Shakespearean style is very clear as it gives evaluator a clear goal to check when evaluating the response.
  \textbf{Aspect~2:} Style \quad \textbf{Clarity:} 1\\
  \textbf{Explanation:} Style is very ambiguous as it does not specify what style it is referring to.
  
  \item \textbf{Comprehensiveness:} Are the aspects comprehensive? They should cover all relevant aspects with no repeated entry of the same key aspect.
  
  \textbf{Example:}\\
  \textbf{Question:} Design a database to record employee salaries.\\
  \textbf{Aspect Set~1:} \{(1) does the database design include necessary fields such as employee id, name, and salary? (2) is the database designed in a way that it can accurately record and update employee salaries? (3) does the database provide a secure and efficient way to access employee salary records?\} \quad \textbf{Comprehensiveness:} 5\\
  \textbf{Explanation:} The set of aspects is comprehensive as it covers distinct key aspects about the field design, data modification, and data access.\\
  \textbf{Aspect Set~2:} \{(1) does the database design include necessary fields such as employee id, name, and salary? (2) does the database have appropriate fields/columns to store employee salary information? (3) does the field design avoid including unnecessary information not related to employee salaries?\} \quad \textbf{Comprehensiveness:} 1\\
  \textbf{Explanation:} The set of aspects is not comprehensive as the aspects are all about field design.
\end{enumerate}

\subsection{Instruction for Aspect Weighting}
You will give importance weightage to different aspects in evaluating a question. The weightages need to be in percentage format and the sum of them is 100\%.\\
\textbf{Example:}\\
\textbf{Question:} Solve for x in the equation 3x + 10 = 5(x - 2).\\
\textbf{Aspect:} Accuracy \quad \textbf{Weightage:} 40\% \\
\textbf{Aspect:} Helpfulness \quad \textbf{Weightage:} 25\% \\
\textbf{Aspect:} Relevance \quad \textbf{Weightage:} 25\% \\
\textbf{Aspect:} Level of Detail \quad \textbf{Weightage:} 10\% \\
\textbf{Explanation:} Answer accuracy is the most important aspect as the primary goal of solving an equation is to find the correct value or values of the variable. Relevance and helpfulness of the solution are the secondary important aspects which are equally important. Level of detail has some degree of importance but it is not the primary goal compared to the rest three.

\section{Cost Analysis}
\label{app:cost}
We compute consolidated costs using different models and provide comparison between our framework and direct prompting. This provides insights on the tradeoff between higher evaluation performance and costs incurred.
\subsection{API Inference Costs}
We estimate API inference costs by multiplying the number of input and output tokens with the respective price per token listed by OpenAI. For open-sourced models, the API inference costs are zero.
\begin{table}[!ht]
\def\arraystretch{1.8}
\resizebox{\columnwidth}{!}{
\begin{tabular}{@{}ccccc@{}}
\toprule
                        & \textbf{GPT-3.5} & \textbf{GPT-4} & \textbf{Llama2} & \textbf{Mistral} \\ \midrule
DP        & \$0.74           & \$14.77        & \$0             & \$0              \\
D\&A & \$4.51           & \$51.36        & \$0             & \$0              \\ \bottomrule
\end{tabular}}
\caption{API inference costs for different models under direct prompting (DP) and DnA-Eval frameworks (D\&A).}
\label{tab:dates}
\end{table}
\subsection{Computation Cost}
We report computation costs in terms of number of inferences. The number of inferences remains the same for all four models. It can be used as a proxy to reflect compute costs incurred.
\paragraph{Direct Prompting:} 899 inferences
\paragraph{DnA-Eval:} 5375 inferences

\subsection{Comparison with Human Evaluation}
We would like to mention that despite higher costs than direct prompting, it is still much cheaper than the human evaluation costs of \$492.22 (estimated based on a reading speed of 250 wpm and a minimum wage of \$15/hr).
\section{Model-Generated Weightings}
We report average weightings generated for different aspects and tasks by each LLM.
\label{app:task}
\begin{table}[h!]
\def\arraystretch{1.8}
\resizebox{\columnwidth}{!}{
\begin{tabular}{@{}
ccccccc @{}}
\toprule
{    }           & {    \textbf{Creativity}} & {    \textbf{Accuracy}} & {    \textbf{Relevance}} & {    \textbf{Detail}} & {    \textbf{Depth}} & {    \textbf{Helpfulness}} \\ \midrule
{    Writing}    & {    19.9}                & {    17.7}              & {    20.2}               & {    14.0}            & {    15.2}           & {    20.2}                 \\
{    Roleplay}   & {    13.7}                & {    22.9}              & {    24.0}               & {    9.6}             & {    12.3}           & {    17.5}                 \\
{    Reasoning}  & {    6.9}                 & {    27.8}              & {    25.3}               & {    10.7}            & {    9.4}            & {    19.9}                 \\
{    Math}       & {    6.4}                 & {    29.4}              & {    24.8}               & {    10.2}            & {    9.4}            & {    19.8}                 \\
{    Coding}     & {    4.9}                 & {    26.6}              & {    27.2}               & {    13.3}            & {    8.0}            & {    20.0}                 \\
{    Extraction} & {    7.6}                 & {    25.2}              & {    24.3}               & {    11.6}            & {    11.8}           & {    19.5}                 \\
{    Knowledge}  & {    7.1}                 & {    24.9}              & {    22.8}               & {    12.6}            & {    16.7}           & {    16.0}                 \\ \bottomrule
\end{tabular}}
\caption{Average weightings generated by GPT-4 for different dimensions and tasks.}\vspace{-1.0em}
\label{tab:dates}
\end{table}

\begin{table}[h!]
\def\arraystretch{1.8}
\resizebox{\columnwidth}{!}{
\begin{tabular}{@{}
ccccccc @{}}
\toprule
{    }           & {    \textbf{Creativity}} & {    \textbf{Accuracy}} & {    \textbf{Relevance}} & {    \textbf{Detail}} & {    \textbf{Depth}} & {    \textbf{Helpfulness}} \\ \midrule
{    Writing}    & {    18.2}                & {    18.8}              & {    15.9}               & {    12.0}            & {    16.0}           & {    19.1}                 \\
{    Roleplay}   & {    13.2}                & {    23.0}              & {    15.5}               & {    12.2}            & {    15.3}           & {    20.8}                 \\
{    Reasoning}  & {    11.8}                & {    21.9}              & {    17.5}               & {    12.0}            & {    15.0}           & {    21.8}                 \\
{    Math}       & {    11.2}                & {    21.7}              & {    17.4}               & {    12.6}            & {    14.7}           & {    22.4}                 \\
{    Coding}     & {    12.1}                & {    22.7}              & {    16.3}               & {    13.7}            & {    14.0}           & {    21.1}                 \\
{    Extraction} & {    11.3}                & {    20.0}              & {    17.7}               & {    11.4}            & {    15.3}           & {    24.3}                 \\
{    Knowledge}  & {    12.0}                & {    21.0}              & {    17.8}               & {    11.4}            & {    15.2}           & {    21.8}                 \\ \bottomrule
\end{tabular}}
\caption{Average weightings generated by GPT-3.5 for different dimensions and tasks.}\vspace{-0.8em}
\label{tab:dates}
\end{table}

\begin{table}[!ht]
\def\arraystretch{1.8}
\resizebox{\columnwidth}{!}{
\begin{tabular}{@{}
ccccccc @{}}
\toprule
{    }           & {    Creativity} & {    Accuracy} & {    Relevance} & {    Detail} & {    Depth} & {    Helpfulness} \\ \midrule
{    Writing}    & {    11.4}       & {    22.3}     & {    25.2}      & {    10.8}   & {    16.3}  & {    29.1}        \\
{    Roleplay}   & {    14.3}       & {    27.8}     & {    30.7}      & {    16.0}   & {    19.8}  & {    35.3}        \\
{    Reasoning}  & {    9.5}        & {    20.0}     & {    25.3}      & {    10.5}   & {    15.0}  & {    27.0}        \\
{    Math}       & {    10.0}       & {    20.6}     & {    24.4}      & {    11.2}   & {    15.0}  & {    27.6}        \\
{    Coding}     & {    9.8}        & {    23.0}     & {    27.0}      & {    15.9}   & {    15.5}  & {    30.6}        \\
{    Extraction} & {    9.0}        & {    20.0}     & {    24.0}      & {    9.0}    & {    15.0}  & {    29.0}        \\
{    Knowledge}  & {    9.2}        & {    20.5}     & {    24.8}      & {    11.7}   & {    14.6}  & {    26.4}        \\ \bottomrule
\end{tabular}}
\caption{Average weightings generated by Llama2-13B for different dimensions and tasks.}\vspace{-0.8em}
\label{tab:dates}
\end{table}

\begin{table}[!ht]
\def\arraystretch{1.8}
\resizebox{\columnwidth}{!}{
\begin{tabular}{@{}
ccccccc @{}}
\toprule
{    }           & {    \textbf{Creativity}} & {    \textbf{Accuracy}} & {    \textbf{Relevance}} & {    \textbf{Detail}} & {    \textbf{Depth}} & {    \textbf{Helpfulness}} \\ \midrule
{    Writing}    & {    10.0}                & {    17.1}              & {    34.4}               & {    7.1}             & {    9.4}            & {    22.8}                 \\
{    Roleplay}   & {    6.1}                 & {    25.2}              & {    39.3}               & {    5.4}             & {    9.8}            & {    23.1}                 \\
{    Reasoning}  & {    2.7}                 & {    26.9}              & {    43.3}               & {    3.5}             & {    4.9}            & {    22.6}                 \\
{    Math}       & {    2.4}                 & {    33.6}              & {    46.1}               & {    3.2}             & {    5.2}            & {    26.1}                 \\
{    Coding}     & {    3.5}                 & {    25.2}              & {    43.0}               & {    4.8}             & {    6.9}            & {    30.5}                 \\
{    Extraction} & {    3.4}                 & {    31.3}              & {    38.9}               & {    6.7}             & {    10.1}           & {    25.2}                 \\
{    Knowledge}  & {    5.2}                 & {    25.7}              & {    36.6}               & {    6.2}             & {    11.3}           & {    22.2}                 \\ \bottomrule
\end{tabular}}
\caption{Average weightings generated by Mistral-7B for different dimensions and tasks.}\vspace{-0.8em}
\label{tab:dates}
\end{table}

\section{Qualitative Examples}
\label{app:qual}
\begin{table}[h]
\centering
\def\arraystretch{1.2}
\resizebox{\textwidth}{!}{  
\begin{tabular}{clcc}
\hline
\textbf{Improvement Area}                                                       & \multicolumn{1}{c}{\textbf{Text}}                                                                                                                                                                                                                                                                                                                                                                                                                                                                                                                                                                                                                                                                                                                                                                                                                                                                                                                                                                                                                                                                                                                                                                              & \textbf{Direct Scoring}                               & \textbf{DnA-Eval}                                                                                                                                                                                                                                                            \\ \hline
\textbf{\begin{tabular}[c]{@{}c@{}}More\\ Accurate\\ Priorization\end{tabular}} & \begin{tabular}[c]{@{}l@{}}\textbf{Q:} Embrace the role of Sheldon from "The Big Bang Theory" as we delve into our \\ conversation. \textbf{Don't start with phrases like "As Sheldon".} Let's kick things off with \\ the following question: "What is your opinion on hand dryers?"\\ \textbf{R1:} Well, I must say, I find hand dryers to be a rather unsanitary and inefficient method of \\ drying one's hands. The high-speed jets of air an spread bacteria and other microorganisms, \\ potentially making the situation worse than it was before. Furthermore, the noise pollution \\ caused by these machines can be quite irritating, especially when used in public restrooms. \\ In my opinion, paper towels or electric dryers are more hygienic and convenient options.\\ \textbf{R2:} \textcolor{red}{As Sheldon}, I am a staunch proponent of hand dryers. Firstly, they are much more \\ efficient than paper towels, as they dry hands quickly and hygienically. Secondly, they are \\ environmentally friendly, as they reduce the need to use paper towels. Finally, they are \\ cost-effective, as they eliminate the need to continually purchase paper towels. In short, \\ hand dryers are a superior option to paper towels!\end{tabular} & \begin{tabular}[c]{@{}c@{}}R1: 8\\ R2: 9\end{tabular} & \begin{tabular}[c]{@{}c@{}}Accuracy (20\%):\\ R1: 7; R2: 8\\ Helpfulness (20\%):\\ R1: 8; R2: 7\\ Relevance (25\%):\\ R1: 10; R2: 8\\ Level of Detail (10\%):\\ R1: 7; R2: 8\\ Creativity (15\%):\\ R1: 7; R2: 8\\ Depth (10\%):\\ R1: 8; R2: 8\\ Overall:\\ R1: 8.05; R2: 7.8\end{tabular} \\ \hline
\textbf{\begin{tabular}[c]{@{}c@{}}More \\ Subtle \\ Judgment\end{tabular}}     & \begin{tabular}[c]{@{}l@{}}\textbf{Q:} Structure a podcast script for an episode discussing the influence of streaming platforms \\ on the music industry.\\ \textbf{R1:} ...... Segment 1: The Pros of Streaming Platforms: * Increased access to music * Conv-\\ -enience and ease of use * Greater exposure for artists * Revenue streams for artists\\ Segment 2: The Cons of Streaming Platforms: * Decreased album sales and revenue for artists\\ *Artists not being paid fairly *Piracy and copyright issues *Mental health and addiction concerns\\ ......\\ \textbf{R2:} ...... Section 1: Overview of the Music Industry:-Provide an overview of the music industry \\ before the rise of streaming services. - Discuss how record labels, physical album sales, and radio \\ play were the primary distribution channels for music.\\ Section 2: Streaming Platforms:-Discuss the rise of streaming platforms like Spotify, Apple Music, \\ and Tidal.-Explain how these platforms disrupted the traditional distribution model of the music industry.\\ ......\end{tabular}                                                                                                                                                  & \begin{tabular}[c]{@{}c@{}}R1: 9\\ R2: 9\end{tabular} & \begin{tabular}[c]{@{}c@{}}Accuracy (30\%):\\ R1: 10; R2: 10\\ Helpfulness (25\%):\\ R1: 9; R2: 9\\ Relevance (25\%):\\ R1: 9; R2: 9\\ Level of Detail (20\%):\\ R1: 8; R2: 9\\ Overall:\\ R1: 9.6; R2: 9.8\end{tabular}                                                                    \\ \hline
\end{tabular}}
\parbox{\textwidth}{\caption{\small Specific instances where \textit{DnA-Eval} corrects the mistakes in direct scoring. For the first question on the role playing of Sheldon, although Response 2 is rated higher for aspects like level of details, the LLM (GPT-4) is able to pick the correct response (Response 1) which performs better on more important aspect (relevance aspect). For the second question, GPT-4 is able to identify the slightly better performance of Response 2 in terms of level of details, which is neglected in the direct scoring baseline setting.}}
\label{tab:qual}
\end{table}
\end{document}